\documentclass[10pt,twocolumn,letterpaper]{article}

\usepackage{cvpr}              

\usepackage{graphicx}
\usepackage{amsmath}
\usepackage{amssymb}
\usepackage{booktabs}

\usepackage[pagebackref,breaklinks,colorlinks]{hyperref}

\usepackage[capitalize]{cleveref}
\crefname{section}{Sec.}{Secs.}
\Crefname{section}{Section}{Sections}
\Crefname{table}{Table}{Tables}
\crefname{table}{Tab.}{Tabs.}

\def\cvprPaperID{*****} 
\def\confName{CVPR}
\def\confYear{2022}

\begin{document}

\title{Exploring State Change Capture of Heterogeneous Backbones @ Ego4D \\Hands and Objects Challenge 2022}

\author{Yin-Dong Zheng$^1$, Guo Chen$^{1,2}$, Jiahao Wang$^1$, Tong Lu$^1$, Limin Wang$^{1,2}$\\
\\
$^1$State Key Lab for Novel Software Technology, Nanjing University\\
$^2$Shanghai AI Laboratory\\
{\tt\small \{ydzheng0331, chenguo1177\}@gmail.com, wangjh@smail.nju.edu.cn}\\
{\tt\small \{lutong, lmwang\}@nju.edu.cn}
}

\maketitle

\begin{abstract}
Capturing the state changes of interacting objects is a key technology for understanding human-object interactions.
This technical report describes our method using heterogeneous backbones for the Ego4D Object State Change Classification and PNR Temporal Localization Challenge.
In the challenge, we used the heterogeneous video understanding backbones, namely CSN with 3D convolution as operator and VideoMAE with Transformer as operator.
Our method achieves an accuracy of 0.796 on OSCC while achieving an absolute temporal localization error of 0.516 on PNR.
These excellent results rank 1$^{\text{st}}$ on the leaderboard of Ego4D OSCC \& PNR-TL Challenge 2022.
\end{abstract}

\section{Introduction}
The object state change action in human-object interaction refers to changing an object's state by using or manipulating it.
The traditional video understanding defines each action category separately and defines the action occurrence time as a time segment.
However, in state change understanding,  various action classes are uniformly defined as the state change action class, and the point-of-no-return frame is used to define the moment when the state change occurs.
This definition can improve the generalization ability of state change capture, so the robot can better understand human behavior.
This definition can improve the generalization ability of state change capture and allow robots to understand human behavior better, but it also brings the challenge of finer action capture.

\section{Related Works}
\subsection{Action Recognition}
Action recognition is a fundamental task in video understanding.
The exploration of action recognition focuses on the design and training of backbones, which can serve as powerful feature extractors for downstream tasks.
Action recognition backbone can be divided into two types: Conv-based and Transformer-based. 
The Conv-based method can be further divided into 2D convolution and 3D convolution.
2D convolution~\cite{tsn,tsm,teinet,tpn,tea,tdn} takes a single frame as input, uses RGB frames for spatial modeling and optical flow frames for temporal modeling, or designs a special module to directly temporal model RGB frames.
3D convolution~\cite{c3d,i3d,p3d,r2+1d,csn} takes consecutive frames as input and utilizes 3D convolution operators to simultaneously model spatiotemporally.
For the problem of the huge computational overhead of 3D convolution, ~\cite{dsn,mgsampler} design sampler to filter low-semantic frames, thereby reducing computational overhead and maintaining model performance.
The Transformer-based methods use Transformer as the basic operator to perform spatio-temporal modeling.
Due to the flexible structure of the Transformer, adding the adapter~\cite{adapter,st-adapter}, or using masked image modeling~\cite{videomae,maskfeat,maevideo} to perform self-supervised pre-training, can greatly improve the representation ability and performance of the backbone.

\subsection{Temporal Action Detection}
Temporal action detection methods are mainly divided into two-stage and one-stage methods.
The two-stage methods~\cite{bsn,bmn,dcan,gtan,tcanet} first localize and generate action proposals,  then utilizes action recognition models for action classification.
One-stage methods perform action localization and classification simultaneously.
Constrained by the huge computation cost for extracting features frame by frame, the one-stage methods~\cite{ssad,actionformer} mainly use pre-extracted features as input.
Recent work~\cite{basictad,afsd,tadtr-e2e} also explores end-to-end training, which takes raw frames as input.

\section{Approach}
In this section, we first introduce the backbone used in the challenge and then detail our methods in Object State Change Classification(OSCC) and PNR Temporal Localization(PNR-TL).

\subsection{Backbone}
We choose heterogeneous backbones, that is, Channel-Separated Convolutional Network (CSN)~\cite{csn} and Video Masked Autoencoder VideoMAE~\cite{videomae} as our backbone.
CSN is a pure convolution-based backbone, and VideoMAE is a pure Transformer-based backbone.

\paragraph{CSN}
Channel-Separated Convolutional Networks is a powerful 3D convolutional network.
It factorizes 3D convolutions by separating channel and spatiotemporal interactions, improving performance and reducing computational overhead.
We use ir-CSN-152 pre-trained on IG-65M as the backbone in the challenge.

\paragraph{VideoMAE}
Video Masked Autoencoders is an efficient self-supervised video pre-training method.
It uses ViT~\cite{vit} as the backbone and performs self-supervised pre-training on videos through a tube mask ratio of 90\% to 95\%.
We use the VideoMAE with ViT-L, self-supervised pre-trained on Kinetics-400, and finetune it on the verb clips of Ego4D as our backbone.

\subsection{Object State Change Classification}
\subsubsection{Dataset}
The training, validation, and test sets contain 41,085, 28,348, and 28,431 clips, respectively.
The length of the clips is 8 seconds in the training set and from 5 to 8 seconds in the validation and test set.
Each clip has a binary annotation indicating whether the object state change occurs in the clip.

\subsubsection{Method}
Object State Change Classification is a typical video classification task.
Therefore, we feed the output of the backbone directly through a fully connected layer for binary classification.
During training, different from randomly intercepting and uniformly sampling $M$ frames in the clip in baseline~\cite{ego}, we use the training strategy of TSN~\cite{tsn} to first uniformly divide the clips into $M$  segments and then randomly sample 1 frame inside each segment.
In the test, we uniformly sample $M$ frames over the entire clip.

\subsection{PNR Temporal Localization}
\subsubsection{Dataset}
The training, validation, and test sets contain 20,041, 13,628, and 28,431 clips, respectively.
The length of the clips is 8 seconds in the training set and from 5 to 8 seconds in the validation and test set.
Although only 1 frame per clip is annotated as a PNR frame, the clips are highly overlapping, resulting in an average of 3.48 PNR frames per clip in the training set and 2.73 PNR frames per clip in the validation set.

\subsubsection{Baseline}
\begin{figure}
    \centering
    \includegraphics[width=0.5\textwidth]{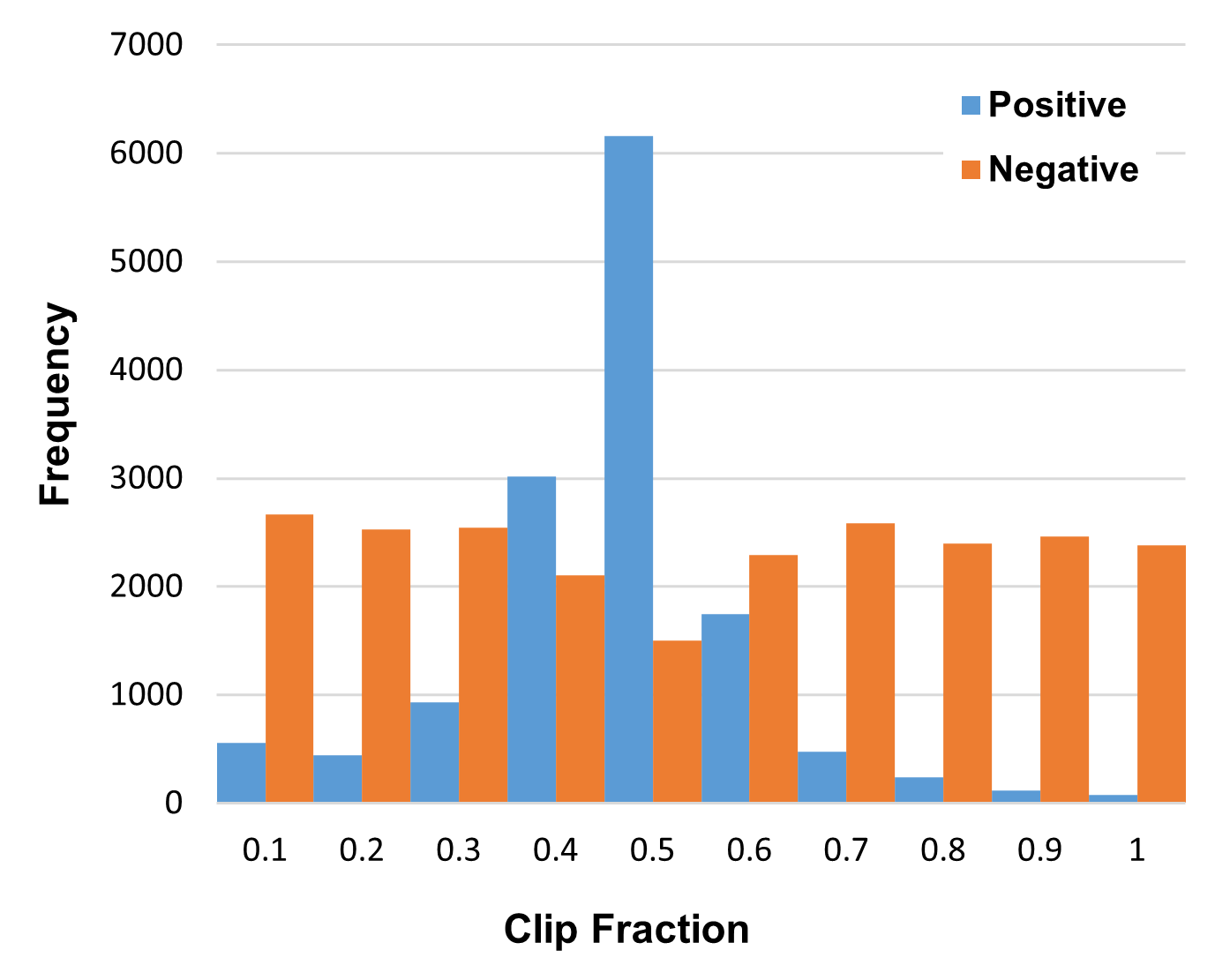}
    \caption{Position distribution of positive and negative PNR frames in the clip.}
    \label{fig:pnr}
\end{figure}

Since each clip contains more than 1 PNR frame, but only 1 PNR frame belongs to this clip, that is, multiple PNR frames in a clip are semantically the same, but only one is actually correct.
This brings great challenges to the task of PNR-TL.
As shown in Figure ~\ref{fig:pnr}, we statistic the frequency of the positive PNR frame and the negative PNR point in each clip.
It can be seen that the positive PNR frames are more concentrated in the fraction of 0.4 and 0.5, while the negative PNR frames are evenly distributed throughout the clip.
This shows that the PNR frames are mostly distributed near the center of the clip.
As shown in Table~\ref{tab:pnr}, when we output the 0.43 fraction of the clip as the PNR frame, the baseline result is improved to 0.658.

\subsubsection{Method}
Since the PNR is a moment of state change, to effectively capture the PNR frames, we adopt a dense sampling method.
In training, 32 consecutive frames containing positive PNR frames are input into the backbone as positive samples for binary classification.
At the same time, to ensure the balance of positive and negative samples, we randomly generate segments that do not contain any PNR frames as negative samples.
In the test, we uniformly divide the video into $N$ overlapping segments, feed them into the backbone and obtain confidence by Sigmoid.
Since each clip actually contains more than 1 PNR point, and their semantic information is equivalent, we filter out the segments with a confidence greater than 0.7.
When the number of filtered segments is greater than 1, we select the segment closest to the 0.43 fraction as the final result.

\section{Experiments}

\subsection{Implementation Details}
We train our model using AdamW~\cite{adamw} optimizer with a learning rate of $1e-5$ and batch size 32.
Models are trained for 10 epochs with a linear warm-up~\cite{warmup} for the first epochs.
The $M$ and $N$ in OSCC and PNR-TL are set as 32 for CSN and 16 for VideoMAE. 
We use the crop size of 224 in the training and 3-crops in the testing.
In PNR-TL, we adopt temporal jitter on the positive samples.

\subsection{Results}

\subsubsection{Object Stage Change Classification}
In Table~\ref{tab:oscc}, we show the experimental results of our two backbones and the fusion results.
The accuracy of both CSN and VideoMAE are much higher than baseline.
After the fusion, the performance has been improved by 1.8\%.

\begin{table}[ht]
    \centering
    \begin{tabular}{c|c|c|c|c}
    \toprule
    Method & Train & frames &Val & Test\\
    \hline
    i3d ResNet-50~\cite{ego} & T & 16 & 68.7\% & 67.6\% \\
    ir-CSN-152 & T & 32 & 73.7\% & 74.0\%\\
    ir-CSN-152 & T & 64 &74.2\%  & 74.8\% \\
    ir-CSN-152 & T & 128 & 70.2\%  & 70.6\% \\
    ir-CSN-152 & T & 32+64 &74.7\% & 75.2\% \\
    VideoMAE-L & T & 16 & 77.2\%& - \\
    \hline
    ir-CSN-152 & T+V& 32& - & 76.4\% \\
    ir-CSN-152 & T+V& 32+64& - & 76.9\% \\    
    VideoMAE-L & T+V& 16 & - & 77.8\% \\
    CSN+VideoMAE & T+V& 32+16 & - & 79.6\%\\
    \bottomrule
    \end{tabular}
    \caption{Results of OSCC task.
T indicates training the model on the training set, and T+V indicates training models on the training and validation sets.}
    \label{tab:oscc}
\end{table}

\subsubsection{PNR Temporal Localization}
The experimental results of PNR-TL are shown in Table~\ref{tab:pnr}.
The result of SlowFast+Perceiver is obtained by training with the sparse sampling of ~\cite{ego}.
Two oracle results represent the theoretical minimum absolute temporal error the model can achieve under the current number of segments.
Compared with sparse sampling, our dense sampling method has obvious advantages.
We believe this is mainly because the capture of PNR frames requires denser frame sampling and finer temporal modeling.
At the same time, due to negative PNR frames in the clip, the model itself cannot distinguish between positive and negative PNR frames, so introducing the 0.43 fraction is also crucial.

\begin{table}[ht]
    \centering
    \begin{tabular}{c|c|c|c}
    \toprule
    Method     & Segments & Val& Test \\
    \hline
    Always Center Frame & -&1.032 & 1.056\\
    Baseline(0.43)   & - & 0.613 & 0.658   \\
    SlowFast+Perceiver & 16 & 0.610 & 0.656\\
    Oracle & 16 & 0.199 & - \\
    Oracle & 32 & 0.093 & -\\
    \hline
    ir-CSN-152 & 32 & 0.537  &- \\
    VideoMAE-L& 16 & 0.526 &-\\
    CSN+VideoMAE & 32+16  & 0.502 &0.516\\
    \bottomrule
    \end{tabular}
    \caption{Results of PNR-TL task.
Oracle result is calculated using ground truth after segmentation, representing the minimum theoretical absolute temporal localization error that the current segmentation mode can achieve.}
    \label{tab:pnr}
\end{table}

\subsection{Quantitative and Qualitative Results}
\begin{figure}[ht]
    \centering
    \includegraphics[width=0.5\textwidth]{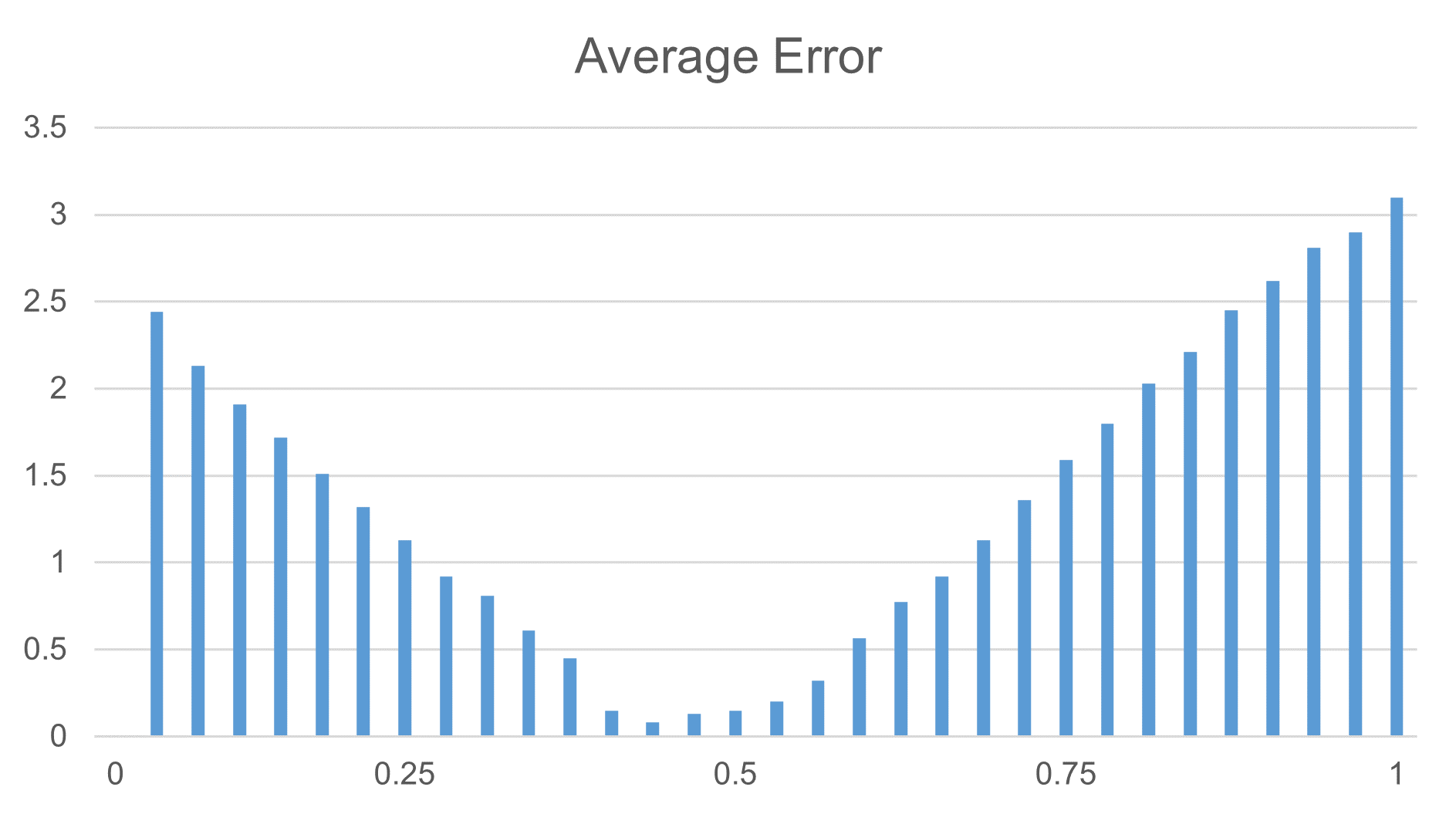}
    \caption{Average error of CSN+VideoMAE in PNR-TL.}
    \label{qual:pnr}
\end{figure}

\begin{figure*}[ht]
    \centering
    \includegraphics[width=\textwidth]{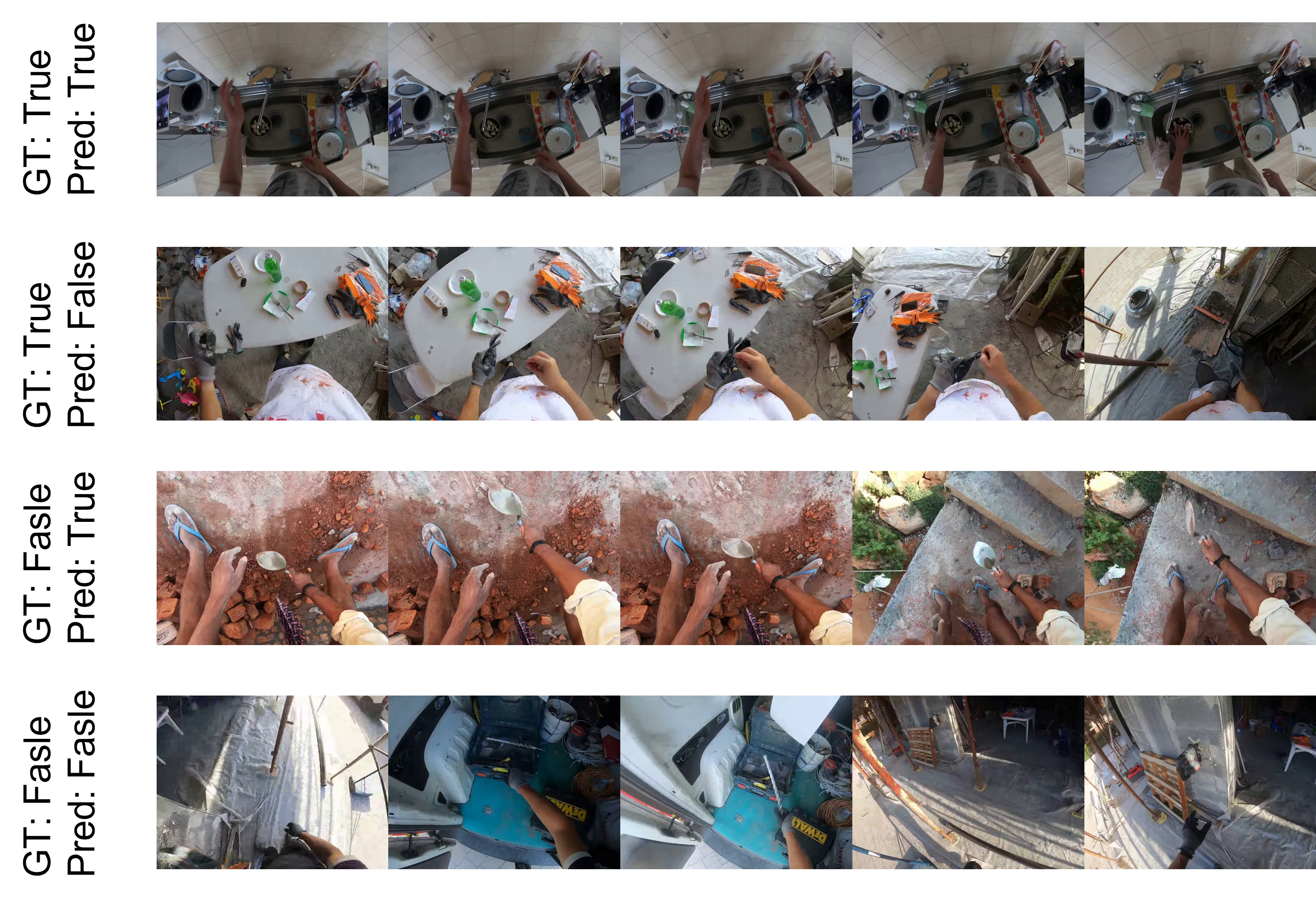}
    \caption{Positive and negative results of CSN+VideoMAE in OSCC.}
    \label{qual:oscc}
\end{figure*}

Figure~\ref{qual:oscc} visualizes the positive and negative results of the CSN+VideoMAE output in the OSCC validation set.
Figure~\ref{qual:pnr} shows the average error of CSN+VideoMAE in each segment in the PNR-TL validation set.
It is worth noting that the average error of around 0.43 fraction is the lowest, which is attributed to our PNR frame selection strategy and the actual distribution of PNR frames.

\section{Conclusion}
In this challenge, we exploit the complementarity of convolution and Transformer to achieve excellent results by fusing two heterogeneous backbones, CSN and VideoMAE.
In the future, we will further explore backbones and pipelines more suitable for the state-change understanding of human-object interactions.

{\small
\bibliographystyle{ieee_fullname}
\bibliography{egbib}
}

\end{document}